\documentclass[journal,onecolumn]{IEEEtran}

\usepackage{graphicx}
\usepackage{booktabs}
\usepackage{algpseudocode}
\usepackage{algorithm}
\usepackage{amssymb}
\usepackage{amsmath}
\usepackage{multirow}
\usepackage{wrapfig}
\usepackage[table]{xcolor}
\usepackage{natbib}
\usepackage{pifont}
\usepackage[font=footnotesize,labelfont=bf]{caption}
\usepackage{enumitem}

\begin{document}

\title{When Robots Sleep: Offline Skill Consolidation for Shared-Policy Robot Learning}

\author{
Nethmi Jayasinghe$^{1}$,
Diana Gontero$^{1}$,
Amit Ranjan~Trivedi$^{1}$%
\thanks{$^{1}$Department of Electrical and Computer Engineering,
University of Illinois at Chicago, Chicago, IL, USA.}%

\thanks{Correspondence: Nethmi Jayasinghe (wjayas3@uic.edu),
Amit Ranjan Trivedi (amitrt@uic.edu).}
}

\maketitle

\begin{abstract}
Robots that learn over long deployments must add new skills without losing the shared policy structure that makes earlier skills reusable. 
We study sequential robot skill learning, where previous trajectories and task losses may be unavailable, and the deployed policy must remain a single shared controller without task-specific heads, routing, or adapters. 
We identify \emph{skill-coupling collapse}, a failure mode in which individual skill success remains non-trivial while reliability among related skills deteriorates. 
We propose \emph{Sleeping Robots}, a wake-sleep framework that learns each new skill during wake and consolidates the shared policy offline during sleep using compact frozen skill memories: frozen critics with unordered state buffers for reinforcement learning and frozen actor snapshots with unordered observation buffers for imitation learning. 
During sleep, these memories define differentiable surrogate objectives whose gradients are combined through Nash bargaining, with adaptive anchoring and local excitability for stable consolidation. 
On Meta-World MT5, Sleeping Robots improves average success by \(64\%\) and pairwise reliability by \(2.0\times\) over the strongest non-oracle baseline, and on SurgicAI it improves average success and backward transfer relative to continual imitation baselines while remaining competitive on pairwise reliability.
\end{abstract}


\IEEEpeerreviewmaketitle


\section{Introduction}

Robots are moving from fixed-task deployment toward long-lived, adaptive operation. A household robot may personalize to a user's routines, an assistive robot may adapt to changing motor abilities, and a manipulation robot may expand its repertoire as new objects and workflows appear. These settings require sequential skill learning under realistic constraints where earlier environments, rewards, demonstrations, or trajectories may no longer be accessible, yet the deployed robot must remain one coherent controller that can reuse and coordinate prior skills. This requirement goes beyond standard per-task retention. A policy may preserve non-trivial success on individual skills while losing the shared structure that makes related skills reliable together. For example, a manipulation policy may still reach or open in isolation, but fail to preserve the coordinated repertoire needed for reliable multi-step interaction. Continual robot learning should therefore evaluate not only whether individual skills survive, but whether they remain reliable inside one shared policy.

We refer to this failure mode as \emph{skill-coupling collapse} where shared-policy compatibility degrades under sequential learning even when isolated skill success remains nonzero. Unlike catastrophic forgetting, which measures loss of individual task performance, skill-coupling collapse asks whether related skills remain jointly reliable after being incorporated into the same parameterization. Average final success can therefore mask a less usable repertoire. We report average final success, backward transfer, and a Pairwise Reliability Score over pre-registered skill pairs to test whether the policy preserves coordinated skill structure rather than only partially retained behaviours.

Existing continual-learning methods reduce interference, but they do not directly target this shared-repertoire objective. Regularization and gradient-projection methods constrain updates to protect previous tasks~\citep{kirkpatrick2017overcoming,chaudhry2018efficient}; replay methods revisit stored experience~\citep{buzzega2020dark,rolnick2019experience}; and distillation methods preserve outputs from earlier models~\citep{li2017learning,schwarz2018progress}. These methods typically act during new-skill acquisition and are evaluated mainly through per-task retention. Modular methods use task-specific heads, routing, adapters, or experts~\citep{wan2024lotus,meng2025preserving,malagon2025self}, but shift part of coordination to module selection rather than maintaining one shared policy at inference. Multi-task gradient methods such as PCGrad and Nash-MTL coordinate objectives when all task losses are available~\citep{yu2020gradient,navon2022multi}; sequential robot learning lacks this access because old task losses and trajectories may be unavailable after learning.

We propose \emph{Sleeping Robots}, an offline consolidation framework for continual robot skill learning with a single deployed policy. During \emph{wake}, the policy learns the current skill using the substrate appropriate to the domain, such as reinforcement learning for state-based manipulation or imitation learning for vision-conditioned surgical control. At wake end, each skill is stored as a compact frozen memory, i.e., a value critic with an unordered state buffer and normalizer snapshot for reinforcement-learning actors, or an actor snapshot with an unordered observation buffer for imitation-trained policies. These memories do not store full trajectories for replay. During \emph{sleep}, the shared policy is consolidated offline without additional environment interaction. Each memory defines a differentiable surrogate objective for the current policy, and the resulting per-skill gradients are combined through Nash bargaining so that no single memory dominates consolidation. Adaptive sleep anchoring adds a conservative pull toward the previous consolidated checkpoint when needed, while local homeostatic excitability stabilizes feedforward reinforcement-learning actors through activation-dependent gain regulation. The sleep analogy motivates the separation between online acquisition and offline consolidation, but the contribution is an algorithmic mechanism for preserving shared robot skill repertoires.

We evaluate \emph{Sleeping Robots} on two five-skill streams: Meta-World MT5 with SAC policies and critic memories, and SurgicAI with vision-conditioned ACT policies and actor memories. The evaluation tests whether offline consolidation preserves isolated success, backward transfer, and shared-repertoire reliability without task-specific heads, routing, adapters, or trajectory replay. On Meta-World MT5, Sleeping Robots improves final average success by \textbf{+64\%}, pairwise reliability by \textbf{2.0$\times$}, and trajectory-structure preservation over the strongest non-oracle baseline. On SurgicAI, it achieves the best average success and backward transfer among continual methods while remaining competitive on pairwise reliability. Supplementary videos are provided. Our contributions are:

\begin{itemize}[leftmargin=1.2em,itemsep=2pt,topsep=3pt,parsep=0pt]
\item We identify \emph{skill-coupling collapse} as a shared-policy reliability failure mode and evaluate it using pairwise reliability alongside average success and backward transfer.
\item We introduce compact frozen skill memories for offline consolidation across reinforcement-learning and imitation-learning policies without replaying prior trajectories.
\item We adapt Nash-bargained gradient coordination to sequential robot learning by replacing unavailable old-task losses with sleep-time gradients from frozen memories.
\item We validate offline consolidation on state-based manipulation and vision-conditioned surgical imitation, with ablations of Nash consolidation, adaptive anchoring, and local excitability.
\end{itemize}

\begin{figure}[t]
    \centering
    \includegraphics[width=\textwidth]{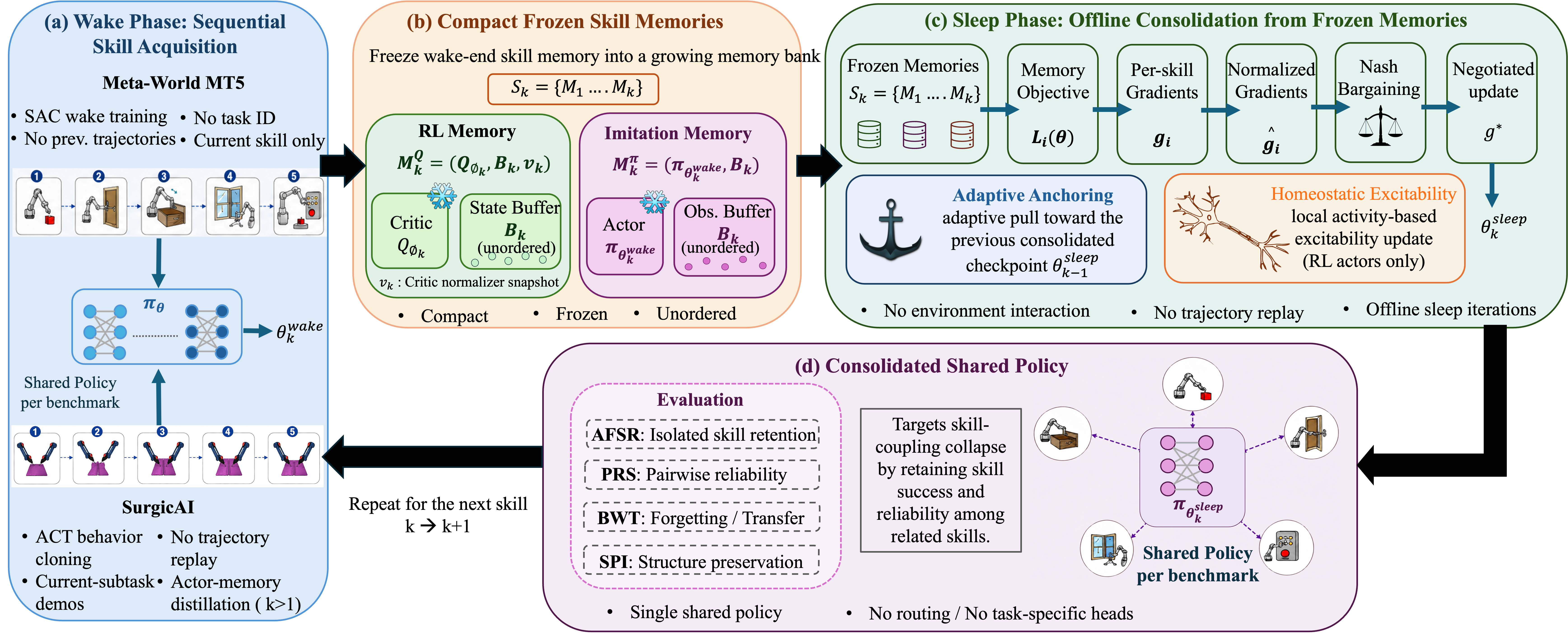}
    \caption{
\textbf{Sleeping Robots overview. }
(a) Wake trains one shared policy on each new skill using only current-skill data. 
(b) The wake-end skill is stored as a compact frozen memory, using critic memories for RL and actor memories for imitation. 
(c) Sleep consolidates the shared policy offline by forming per-skill objectives from frozen memories, normalizing gradients, and combining them with Nash bargaining, with optional anchoring and homeostatic excitability, to obtain \(\theta_k^{\mathrm{sleep}}\) without environment interaction or trajectory replay. 
(d) The consolidated policy is evaluated across learned skills, and the wake-sleep cycle repeats for the next skill.
\vspace{-15pt}
}
    \label{fig:sleeping_robots_overview}
\end{figure}

\section{Related Work}
\label{sec:related}

\noindent\textbf{Continual learning for robot policies.}
Continual-learning methods reduce interference through regularization, projection, replay, or distillation. EWC~\citep{kirkpatrick2017overcoming}, SI~\citep{zenke2017continual}, and MAS~\citep{aljundi2018memory} penalize changes to parameters important for previous tasks, while GEM~\citep{lopez2017gradient} and A-GEM~\citep{chaudhry2018efficient} constrain new-task gradients using episodic memories. Replay methods such as DER~\citep{buzzega2020dark} and CLEAR~\citep{rolnick2019experience} revisit stored examples or transitions, and OWL~\citep{kessler2022same} uses task-specific output heads in continual RL. These methods mainly protect per-task retention during acquisition and often rely on stored examples, transitions, task outputs, or task-specific heads. Our setting instead targets cross-skill reliability in one shared policy, using compact frozen memories and unordered state or observation buffers.

\noindent\textbf{Distillation, consolidation, and modular skills.}
LwF~\citep{li2017learning} preserves prior outputs during new-task training, Progress \& Compress~\citep{schwarz2018progress} separates acquisition from compression, and Reset \& Distill~\citep{ahn2025reset} trains per-task experts before distilling them into a unified actor. Modular and skill-library methods, including PackNet~\citep{mallya2018packnet}, HAT~\citep{serra2018overcoming}, LOTUS~\citep{wan2024lotus}, LEGION~\citep{meng2025preserving}, AtomicVLA~\citep{zhang2026atomicvla}, and CompoNet~\citep{malagon2025self}, reduce interference through task-specific capacity, routing, adapters, or modules. These approaches can preserve individual skills, but they either constrain acquisition online or shift coordination to module selection. Sleeping Robots keeps one policy at inference and adds an offline phase in which frozen skill memories jointly consolidate the shared parameterization. In vision-conditioned imitation, this separates our approach from these prior works.

\noindent\textbf{Gradient coordination under sequential access.}
Multi-task optimization methods address gradient conflict when all task objectives are available. PCGrad~\citep{yu2020gradient} projects conflicting gradients, CAGrad~\citep{liu2021conflict} chooses conflict-averse directions, and Nash-MTL~\citep{navon2022multi} formulates multi-task learning as bargaining over task gradients. Sleeping Robots also coordinates gradients, but in a sequential setting where old environments, demonstrations, trajectories, and losses may no longer be available. We replace direct old-task losses with frozen memories that generate sleep-time gradients, allowing previous skills to participate in consolidation after their original data are gone.

\noindent\textbf{Offline and biologically inspired consolidation.}
Offline consolidation and sleep-like phases have been studied in biologically inspired learning. The wake-sleep algorithm~\citep{hinton1995wake} alternates generative and recognition updates, while later continual-learning work uses generative replay or activity-dependent stabilization to reduce forgetting~\citep{van2020brain,tadros2022sleep}. Sleeping Robots differs in both domain and mechanism: it targets robot policies across reinforcement learning and imitation learning, avoids trajectory replay, and consolidates one shared policy through Nash-bargained optimization over frozen skill memories. Links to sleep, homeostatic plasticity, and shared motor adaptation~\citep{tononi2014sleep,turrigiano2008self,wolpaw2022heksor} serve as design motivation rather than claims of neural fidelity.

\section{Method: Sleeping Robots}
\label{sec:method}

\noindent\textbf{Sequential shared-policy skill learning.}
We consider a robot learning an ordered skill sequence \(\mathcal{T}_{1:K}=\{\mathcal{T}_1,\ldots,\mathcal{T}_K\}\). A single shared policy \(\pi_\theta\) is updated as skills arrive. During skill \(k\), the learner accesses only data or interaction from \(\mathcal{T}_k\), and previous task trajectories are unavailable. The policy must retain a growing repertoire within one parameterization, without task-specific policy heads, routing, or adapters at inference. For reinforcement learning, each skill is an MDP with shared state and action spaces but skill-specific dynamics, rewards, and initial-state distributions. For imitation learning, each skill is specified by demonstrations over observations and expert actions. We denote parameters after wake training on skill \(k\) by \(\theta_k^{\mathrm{wake}}\) and after sleep consolidation by \(\theta_k^{\mathrm{sleep}}\).

\noindent\textbf{Wake phase and frozen skill memories.}
Sleeping Robots separates online skill acquisition from offline consolidation. During \emph{wake}, the policy learns the current skill using the substrate appropriate to the domain. For reinforcement-learning actors, wake uses only current-skill RL updates, stored memories are not consulted, previous trajectories are not replayed, and the policy receives no task identifier. For deterministic imitation-learning actors, wake uses current-skill behavior cloning, with output-space distillation on stored observation buffers when previous actor memories are available. This wake-time distillation uses the same frozen actor memories later used during sleep and introduces no trajectory, reward, or transition replay.

At wake end, the learned skill is stored as a compact frozen memory \(\mathcal{M}_k\) that provides a differentiable sleep objective. For reinforcement-learning actors, we train and freeze a skill critic \(\mathcal{M}_k^{Q}=(Q_{\phi_k},\mathcal{B}_k,\nu_k)\), where \(\mathcal{B}_k\) is an unordered state buffer and \(\nu_k\) is the critic normalizer snapshot. The buffer contains no actions, rewards, or transitions, and sleep evaluates the current policy through the frozen critic rather than replaying past trajectories. For imitation-learning actors, we freeze the wake-end policy and store \(\mathcal{M}_k^{\pi}=(\pi_{\theta_k^{\mathrm{wake}}},\mathcal{B}_k)\), where \(\mathcal{B}_k\) contains unordered observations such as robot states and images. After sleep, \(\theta_k^{\mathrm{sleep}}\) is stored as the consolidated checkpoint for future anchoring. Both substrates expose the same interface, where a frozen memory defines a skill objective while inference uses one shared policy.

\noindent\textbf{Sleep objectives from frozen memories.}
After wake training on skill \(k\), the learner enters \emph{sleep} only when \(k\geq 2\). Sleep uses no environment interaction, performs no memory update, and accesses only the unordered buffers in the stored memory set \(\mathcal{S}_k=\{\mathcal{M}_i\}_{i=1}^{k}\). Each memory defines a differentiable objective \(L_i(\theta)\) for the current shared policy. For reinforcement-learning memories, the frozen critic induces \(L_i^{Q}(\theta)=-\mathbb{E}_{s\sim\mathcal{B}_i}\left[Q_{\phi_i}\bigl(\nu_i(s),\pi_\theta(s)\bigr)\right]\), where \(\nu_i\) is the stored normalizer. For imitation-learning memories, sleep uses frozen-actor distillation \(L_i^{\pi}(\theta)=\mathbb{E}_{o\sim\mathcal{B}_i}\left[\|\pi_\theta(o)-\pi_{\theta_i^{\mathrm{wake}}}(o)\|_2^2\right]\), where \(o\) denotes the stored observation input. The skill-level gradient is \(g_i=\nabla_\theta L_i(\theta)\). Thus, frozen memories provide consolidation signals without replaying old actions, rewards, transitions, or rollouts. For RL memories, the critic-induced objective is evaluated only on stored late-wake states using the normalizer snapshot saved at wake end. We monitor frozen-critic value changes, Nash-gradient norms, and stopping iterations as diagnostics for critic extrapolation.

\noindent\textbf{Nash-bargained sleep consolidation.}
Sleep must combine gradients from multiple frozen skill memories without letting one memory dominate the shared-policy update. A simple mean is sensitive to gradient scale and can move the policy toward high-magnitude memories even when other skills disagree. We therefore normalize each memory gradient as \(\hat{g}_i=g_i/(\|g_i\|_2+\epsilon)\) and choose convex weights \(\alpha^\star\in\Delta^{k-1}\) by solving
\begin{equation}
\alpha^\star =
\arg\max_{\alpha\in\Delta^{k-1}}
\sum_{i=1}^{k}
\log\left(
\left\langle
\hat{g}_i,
\sum_{j=1}^{k}\alpha_j \hat{g}_j
\right\rangle
+
\epsilon_{\mathrm{Nash}}
\right).
\label{eq:nash_objective}
\end{equation}
The resulting sleep direction is \(g^\star=\sum_{i=1}^{k}\alpha_i^\star\hat{g}_i\). The log-product objective favors directions that maintain agreement with all stored skill memories, giving a scale-invariant analogue of Nash-MTL for the sequential setting where old task losses are unavailable. The floor \(\epsilon_{\mathrm{Nash}}\) prevents numerical failure when agreement is weak. We do not require \(g^\star\) to descend every memory objective at every step; instead, it is a negotiated direction that limits domination by any single memory. We compute \(\alpha^\star\) with warm-started Frank-Wolfe iterations and update the policy with \(\theta\leftarrow\mathrm{Adam}(\theta,g^\star,\eta_{\mathrm{sleep}})\) when anchoring is inactive. A fixed sleep budget, gradient normalization, the numerical floor, and early stopping keep offline updates controlled.

\noindent\textbf{Adaptive sleep anchoring.}
Frozen-memory gradients can still move the policy too aggressively. We therefore use a wake-end proximity gate to decide when sleep should remain close to the previous consolidated checkpoint. At the start of sleep after skill \(k\), we compute \(c_i=\|\theta-\theta_i^{\mathrm{wake}}\|_2\) for \(i<k\). This quantity is used only as a conservative anchoring trigger, not as a behavioral shift estimate or memory-reliability certificate. The anchoring coefficient is
\begin{equation}
\lambda_k =
\begin{cases}
\lambda, & \text{if } \min_{i<k} c_i < \tau,\\
0, & \text{otherwise},
\end{cases}
\label{eq:lambda_anchoring}
\end{equation}
where \(\lambda\) is fixed. Sleep then uses the anchored objective \(L_{\mathrm{sleep}}(\theta)=\sum_{i=1}^{k}\alpha_i^\star L_i(\theta)+\frac{\lambda_k}{2}\|\theta-\theta_{k-1}^{\mathrm{sleep}}\|_2^2\), with update \(\theta\leftarrow\mathrm{Adam}(\theta,g^\star+\lambda_k(\theta-\theta_{k-1}^{\mathrm{sleep}}),\eta_{\mathrm{sleep}})\). The anchor is used only during sleep. It does not constrain wake learning, remove any memory from bargaining, or replace the Nash-negotiated update.

\noindent\textbf{Local excitability for feedforward RL actors.}
For feedforward RL actors, Sleeping Robots adds a local excitability variable \(e_i\in[e_{\min},e_{\max}]\) to each hidden neuron, giving activation \(y_i=e_i\sigma(\mathbf{w}_i^\top\mathbf{x})\), where \(\sigma\) is ReLU. Unlike ordinary learnable gains, \(e_i\) is not updated by backpropagation. It changes only through local activation statistics:
\begin{equation}
\bar{a}_i
\leftarrow
(1-\beta)\bar{a}_i + \beta y_i,
\qquad
e_i
\leftarrow
\mathrm{clip}_{[e_{\min},e_{\max}]}
\left(
e_i + \eta_e(\alpha_{\mathrm{target}}-\bar{a}_i)
\right).
\label{eq:excitability_update}
\end{equation}
Here, \(\bar{a}_i\) tracks activity and \(\alpha_{\mathrm{target}}\) is a target activity level. Phase-specific rates \(\eta_e^{\mathrm{wake}}\) and \(\eta_e^{\mathrm{sleep}}\) are used during wake and sleep. Transformer-based imitation policies use the same memory and sleep components without excitability. Thus, in the RL setting, policy parameters are updated by RL during wake and Nash-bargained consolidation during sleep, while excitability changes only through local activity. Algorithm~\ref{alg:sleeping_robots} in the appendix summarizes the wake-sleep lifecycle. Wake learns the current skill and freezes a compact memory. Sleep computes per-skill objectives from all stored memories, combines their gradients through Nash bargaining, and optionally applies adaptive anchoring toward the previous consolidated checkpoint. At inference, Sleeping Robots uses a single shared policy $\pi_\theta$ with no task-specific routing, adapters, or separate policy heads.

\section{Experiments}
\label{sec:experiments}

\noindent\textbf{Benchmarks.}
We evaluate Sleeping Robots on two five-skill streams using the same sleep-time consolidation algorithm with different policy substrates and memory types. \textbf{Meta-World MT5} uses simulated Sawyer manipulation tasks~\citep{yu2020meta} with the stream \textsc{reach}, \textsc{door-open}, \textsc{drawer-open}, \textsc{window-close}, and \textsc{button-press}. Policies are trained with SAC~\citep{haarnoja2018soft}, and each skill memory is a frozen critic with an unordered state buffer. \textbf{SurgicAI} uses a da Vinci-compatible suturing simulator~\citep{wu2024surgicai} with the stream \textsc{approach}, \textsc{place}, \textsc{insert}, \textsc{regrasp}, and \textsc{pullout}. Policies are vision-conditioned ACT models~\citep{zhao2023learning}, and each skill memory is a frozen actor snapshot with an unordered observation buffer. Within each benchmark, all methods use the same skill order, policy substrate, wake-training budget, and evaluation protocol. 

\noindent\textbf{Metrics.}
After training through skill \(k\), we evaluate the shared policy on every learned skill \(i\leq k\) and record the success rate \(S_i^{(k)}\). We report average final success rate,
\begin{equation}
\mathrm{AFSR}_k=\frac{1}{k}\sum_{i=1}^{k}S_i^{(k)},
\label{eq:afsr}
\end{equation}
which measures isolated retention. To measure shared-repertoire reliability, we use a Pairwise Reliability Score over pre-registered related skill pairs \(\mathcal{C}_k\):
\begin{equation}
\mathrm{PRS}_k
=
\frac{1}{|\mathcal{C}_k|}
\sum_{(i,j)\in\mathcal{C}_k}
\min\left(S_i^{(k)},S_j^{(k)}\right).
\label{eq:prs}
\end{equation}
PRS is reported for \(k\geq2\). The pairwise minimum penalizes cases where one skill in a related pair survives but the other collapses, which is the reliability failure targeted by skill-coupling collapse. For MT5, \(\mathcal{C}_k\) contains pre-registered manipulation pairs that share reaching, contact, and object-interaction structure. For SurgicAI, \(\mathcal{C}_k\) contains workflow-ordered pairs reflecting adjacent and skip-step dependencies in the suturing sequence. We also report backward transfer, where negative values indicate forgetting of earlier skills, and a Synergy Preservation Index measuring preservation of the reference-skill trajectory subspace after the full stream. Full definitions of BWT, SPI, pair lists, and memory accounting are provided in Appendices~\ref{app:implementation} and~\ref{app:baseline_details}.

\begin{figure}[t]
    \centering
    \includegraphics[width=\textwidth]{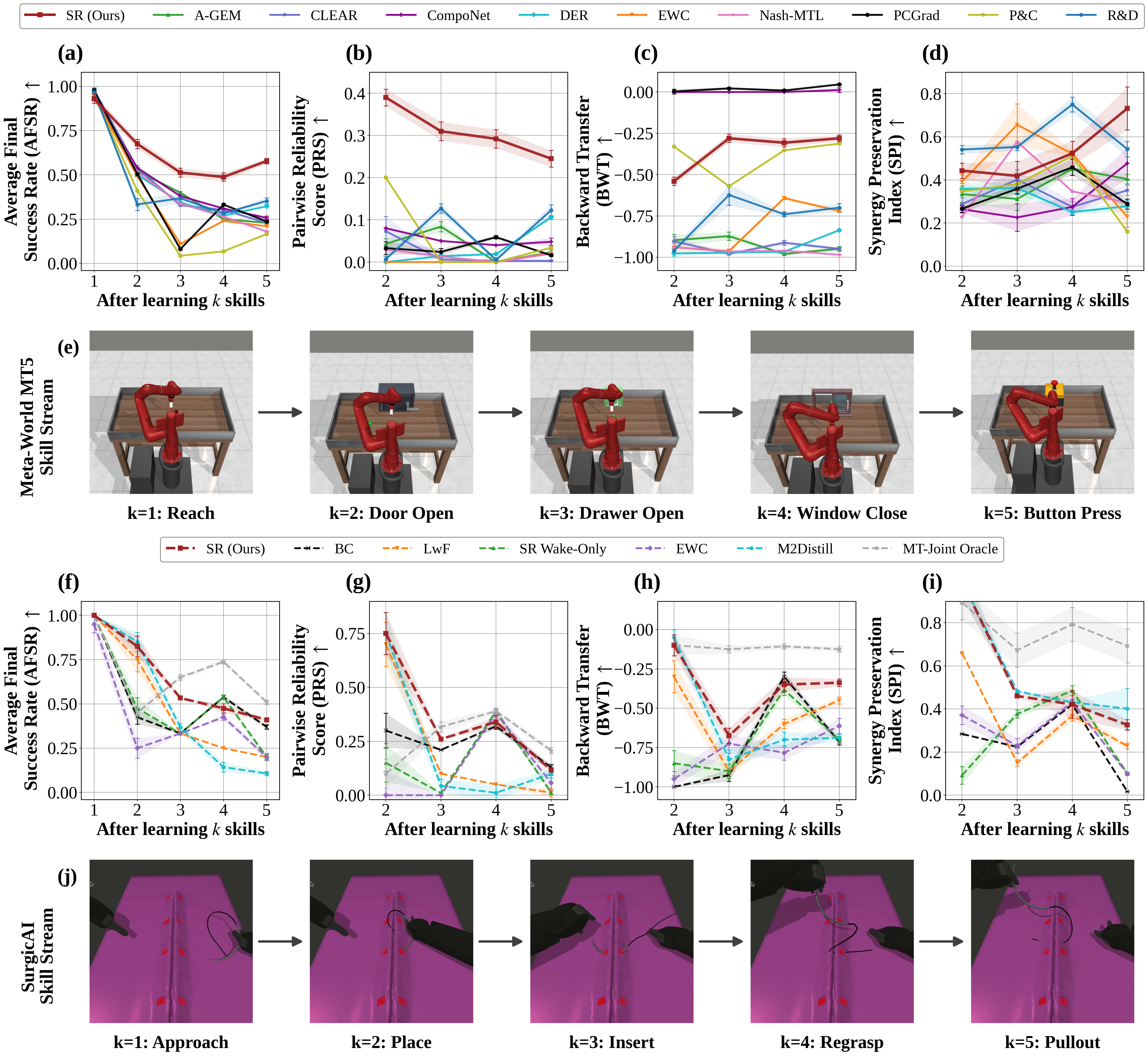}
    \caption{
    Continual learning dynamics on Meta-World MT5 and SurgicAI. Panels (a)--(d) report MT5 AFSR, PRS, BWT, and SPI; panels (f)--(i) report the same metrics for SurgicAI. Panels (e) and (j) show the corresponding five-skill streams. Higher AFSR, PRS, and SPI are better, while less negative BWT indicates less forgetting. Shaded regions denote standard error over three seeds; for SurgicAI, these are evaluation seeds.
    \vspace{-20pt}
    }
    \label{fig:dynamics}
\end{figure}

\begin{table}[h]
\centering
\small
\setlength{\tabcolsep}{5.5pt}
\renewcommand{\arraystretch}{1.08}
\caption{
Final MT5 performance after sequentially learning five Meta-World skills.
Results are mean $\pm$ standard error. Higher is better; bold and underline indicate best and second-best values.
}
\label{tab:mt5_main_results}
\begin{tabular}{p{2.05cm}lcccc}
\toprule
\textbf{Category}
& \textbf{Method}
& \textbf{AFSR$\uparrow$}
& \textbf{PRS$\uparrow$}
& \textbf{BWT$\uparrow$}
& \textbf{SPI$\uparrow$} \\
\midrule

\multirow{2}{=}{Stability-constrained}
& EWC
& 0.210\,$\pm$\,0.001
& 0.021\,$\pm$\,0.000
& $-$0.719\,$\pm$\,0.008
& 0.229\,$\pm$\,0.048 \\

& A-GEM
& 0.229\,$\pm$\,0.007
& 0.023\,$\pm$\,0.006
& $-$0.949\,$\pm$\,0.014
& 0.404\,$\pm$\,0.022 \\

\midrule
\multirow{2}{=}{Replay-based}
& DER
& 0.323\,$\pm$\,0.005
& 0.106\,$\pm$\,0.004
& $-$0.836\,$\pm$\,0.002
& 0.214\,$\pm$\,0.059 \\

& CLEAR
& 0.229\,$\pm$\,0.004
& 0.003\,$\pm$\,0.002
& $-$0.949\,$\pm$\,0.008
& 0.347\,$\pm$\,0.038 \\

\midrule
\multirow{2}{=}{Distillation-based}
& R\&D
& \underline{0.353\,$\pm$\,0.018}
& \underline{0.123\,$\pm$\,0.013}
& $-$0.700\,$\pm$\,0.025
& \underline{0.504\,$\pm$\,0.060} \\

& P\&C
& 0.166\,$\pm$\,0.000
& 0.034\,$\pm$\,0.000
& $-$0.312\,$\pm$\,0.000
& 0.160\,$\pm$\,0.000 \\

\midrule
\multirow{3}{=}{Multi-skill coordination}
& Nash-MTL
& 0.180\,$\pm$\,0.000
& 0.020\,$\pm$\,0.000
& $-$0.985\,$\pm$\,0.000
& 0.295\,$\pm$\,0.000 \\

& PCGrad
& 0.237\,$\pm$\,0.004
& 0.016\,$\pm$\,0.020
& \textbf{+0.046\,$\pm$\,0.005}
& 0.288\,$\pm$\,0.023 \\

& CompoNet
& 0.257\,$\pm$\,0.011
& 0.048\,$\pm$\,0.009
& \underline{+0.012\,$\pm$\,0.014}
& 0.477\,$\pm$\,0.073 \\

\midrule
\textbf{Ours}
& \textbf{Sleeping Robots}
& \textbf{0.578\,$\pm$\,0.016}
& \textbf{0.245\,$\pm$\,0.020}
& $-$0.280\,$\pm$\,0.020
& \textbf{0.767\,$\pm$\,0.017} \\

\bottomrule
\end{tabular}
\end{table}

\begin{table}[h]
\centering
\small
\setlength{\tabcolsep}{5.5pt}
\renewcommand{\arraystretch}{1.08}
\caption{
Final SurgicAI performance after sequentially learning five surgical subtasks.
MT-Joint is an oracle with joint access to all skills, not a continual-learning method.
Results are mean $\pm$ standard error over three evaluation seeds. Higher is better; bold and underline indicate best and second-best values.
}
\label{tab:surgicai_main_results}
\begin{tabular}{p{2.2cm}lcccc}
\toprule
\textbf{Category}
& \textbf{Method}
& \textbf{AFSR$\uparrow$}
& \textbf{PRS$\uparrow$}
& \textbf{BWT$\uparrow$}
& \textbf{SPI$\uparrow$} \\
\midrule

Sequential BC
& BC
& \underline{0.370\,$\pm$\,0.013}
& \textbf{0.129\,$\pm$\,0.014}
& $-$0.713\,$\pm$\,0.020
& 0.017\,$\pm$\,0.010 \\

\midrule
CL baselines
& EWC
& 0.217\,$\pm$\,0.020
& 0.057\,$\pm$\,0.034
& $-$0.613\,$\pm$\,0.048
& 0.099\,$\pm$\,0.007 \\

& LwF
& 0.200\,$\pm$\,0.010
& 0.012\,$\pm$\,0.020
& \underline{$-$0.450\,$\pm$\,0.022}
& 0.228\,$\pm$\,0.014 \\

& M2Distill
& 0.107\,$\pm$\,0.012
& 0.099\,$\pm$\,0.020
& $-$0.688\,$\pm$\,0.024
& \textbf{0.401\,$\pm$\,0.095} \\

\midrule
Ours / ablation
& SR Wake-Only
& 0.200\,$\pm$\,0.015
& 0.012\,$\pm$\,0.010
& $-$0.700\,$\pm$\,0.020
& 0.101\,$\pm$\,0.006 \\

& \textbf{Sleeping Robots}
& \textbf{0.410\,$\pm$\,0.010}
& \underline{0.117\,$\pm$\,0.012}
& \textbf{$-$0.338\,$\pm$\,0.024}
& \underline{0.326\,$\pm$\,0.024} \\

\bottomrule

\rowcolor{gray!10}
Oracle
& MT-Joint
& 0.510\,$\pm$\,0.013
& 0.207\,$\pm$\,0.014
& $-$0.125\,$\pm$\,0.017
& 0.693\,$\pm$\,0.079 \\

\bottomrule
\end{tabular}
\vspace{-15pt}
\end{table}

\noindent\textbf{Implementation details and fairness.}
All methods are evaluated under the same skill order, policy substrate, wake-training budget, and evaluation protocol within each benchmark. During wake, each method trains on the current skill under its specified continual-learning rule. No method receives previous trajectories or additional environment interaction. Sleeping Robots performs sleep consolidation without environment interaction and updates only from stored skill memories. We evaluate after each skill and report final performance after the five-skill stream. MT5 results use three random seeds with \(N=100\) evaluation episodes per skill. SurgicAI uses three evaluation seeds with \(N=20\) episodes per skill due to simulator cost. Full optimizer settings, sleep hyperparameters, memory budgets, baseline configurations, and simulator details are provided in Appendix~\ref{app:implementation}.

\noindent\textbf{Baselines.}
For Meta-World MT5, we compare against representative continual-learning and multi-skill coordination baselines. Stability-constrained methods include EWC~\citep{kirkpatrick2017overcoming} and A-GEM~\citep{chaudhry2018efficient}; replay-based methods include DER~\citep{buzzega2020dark} and CLEAR~\citep{rolnick2019experience}; distillation-based methods include Reset~\&~Distill~\citep{ahn2025reset} and Progress~\&~Compress~\citep{schwarz2018progress}; and coordination methods include Nash-MTL~\citep{navon2022multi}, PCGrad~\citep{yu2020gradient}, and CompoNet~\citep{malagon2025self}. All use the same SAC actor architecture, skill order, training budget, and evaluation protocol as Sleeping Robots. For SurgicAI, all baselines use the same ACT policy substrate and five-skill stream. We compare against sequential behaviour cloning, EWC, LwF~\citep{li2017learning}, and M2Distill~\citep{roy2025m2distill}. We also include SR Wake-Only to isolate the effect of offline sleep consolidation, and MT-Joint as an oracle reference with joint access to all five subtasks rather than as a continual-learning baseline.

\noindent\textbf{Meta-World MT5 results and ablations.}
Figure~\ref{fig:dynamics} shows learning dynamics on the five-skill MT5 stream, and Table~\ref{tab:mt5_main_results} reports final performance. Across baselines, AFSR decreases as new skills arrive, while PRS often drops more sharply, indicating that shared policies can retain partial single-skill competence while losing reliability among related skills. Sleeping Robots achieves the best final AFSR, PRS, and SPI. It reaches AFSR \(0.578\pm0.016\), improving over the strongest non-oracle baseline, Reset~\&~Distill at \(0.353\pm0.018\), by more than \(60\%\) relative. The reliability gain is larger, with PRS \(0.245\pm0.020\), approximately \(2.0\times\) the best baseline PRS of \(0.123\pm0.013\). Sleeping Robots also obtains the highest SPI, \(0.767\pm0.017\), compared with \(0.504\pm0.060\) for the strongest baseline. Although PCGrad has the highest BWT, it has much lower AFSR, PRS, and SPI, showing that backward transfer alone can miss whether the shared policy preserves a reliable skill repertoire.

Table~\ref{tab:mt5_ablations} ablates the main components after the full MT5 stream. Wake-only performs worst, with low AFSR, zero PRS, and strongly negative BWT, confirming that sequential wake learning alone does not preserve pairwise reliability. Mean-gradient sleep improves over wake-only but remains below full Sleeping Robots, supporting the use of scale-normalized Nash negotiation. Removing the anchor lowers AFSR, PRS, and SPI. Removing excitability gives the highest PRS but reduces AFSR, BWT, and especially SPI, while disabling sleep-time excitability gives the best BWT but substantially reduces AFSR and PRS. Full Sleeping Robots gives the strongest overall balance, achieving the best AFSR and SPI with second-best PRS and BWT.

\begin{table}[t]
\centering
\small
\setlength{\tabcolsep}{3.5pt}
\renewcommand{\arraystretch}{1.08}
\caption{
MT5 ablations after learning all five skills ($k{=}5$), using seed 42 and $N{=}100$ evaluation episodes per skill.
Higher is better; bold and underline indicate best and second-best values.
}
\label{tab:mt5_ablations}
\begin{tabular}{lccccccc}
\toprule
\textbf{Metric}
& \textbf{Wake-only}
& \textbf{Mean-grad}
& \textbf{No anchor}
& \textbf{No exc.}
& \textbf{Wake-exc off}
& \textbf{Sleep-exc off}
& \textbf{SR} \\
\midrule
AFSR $\uparrow$
& 0.204 & 0.336 & 0.336 & 0.410 & 0.364 & 0.299 &\textbf{0.578} \\
PRS $\uparrow$
& 0.000 & 0.120 & 0.097 & \textbf{0.272} & 0.150 & 0.044 & \underline{0.245} \\
BWT $\uparrow$
& $-$0.970 & $-$0.802 & $-$0.605 & $-$0.392 & $-$0.688 & \textbf{$-$0.167} & \underline{$-$0.280} \\
SPI $\uparrow$
& 0.342 & 0.426 & 0.296 & 0.233 & 0.479 & 0.492 &\textbf{0.812} \\
\bottomrule
\end{tabular}
\vspace{-15pt}
\end{table}

We further evaluate order and length sensitivity using a reverse MT5 ordering and an eight-skill Meta-World stream, with details in Appendix~\ref{app:order_length}. As shown in Table~\ref{tab:order_length_sensitivity}, Sleeping Robots retains nontrivial final success, pairwise reliability, and trajectory-structure preservation under both variants, suggesting that the MT5 trends are not tied only to the original skill order.

\begin{wrapfigure}[8]{r}{0.50\textwidth} \vspace{-1em} 
\centering \begin{minipage}{0.48\textwidth} 
\footnotesize \centering 
\captionof{table}{Meta-World order and length sensitivity. Final metrics for original MT5, reverse MT5, and MT8.
}\label{tab:order_length_sensitivity} 
\begin{tabular}{lccc} 
\toprule 
\textbf{Metric} & \textbf{MT5} & \textbf{Rev. MT5} & \textbf{MT8} \\ 
\midrule
AFSR $\uparrow$ & $0.578$ & $0.501$ & $0.401$\\ 
PRS $\uparrow$ & $0.245$ & $0.197$ & $0.188$ \\ BWT $\uparrow$ & $-0.280$ & $-0.378$ & $-0.363$\\ 
SPI $\uparrow$ & $0.767$ & $0.659$ & $0.583$ \\ \bottomrule 
\end{tabular} 
\end{minipage}
\vspace{-1.0em} 
\end{wrapfigure}

\noindent\textbf{SurgicAI results.}
We next test whether the same consolidation principle transfers from state-based RL to vision-conditioned surgical imitation. SurgicAI uses ACT policies with frozen actor memories instead of SAC policies with critic memories. Figure~\ref{fig:dynamics} shows learning dynamics, and Table~\ref{tab:surgicai_main_results} reports final performance after the five-subtask stream. Sleeping Robots achieves the highest AFSR among continual methods, \(0.410\pm0.010\), and the best BWT, \(-0.338\pm0.024\). Relative to SR Wake-Only, offline sleep more than doubles AFSR and improves BWT from \(-0.700\pm0.020\) to \(-0.338\pm0.024\), showing that the gain is not explained by wake-time distillation alone. The compatibility metrics show a trade-off. Sequential BC has the highest non-oracle PRS, \(0.129\pm0.014\), while Sleeping Robots is close at \(0.117\pm0.012\). M2Distill has the highest non-oracle SPI, \(0.401\pm0.095\), but much lower AFSR and BWT. Overall, Sleeping Robots gives the strongest retention-reliability balance on SurgicAI where it achieves the best continual-method AFSR and BWT, while remaining close to the best non-oracle PRS, without task-specific heads, routing, or trajectory replay.

\section{Conclusion and Limitations}
\label{sec:conclusion_limitations}

We introduced Sleeping Robots, an offline consolidation framework for continual robot skill learning with a single shared policy. The main finding is that continual robot learning should not be evaluated only through isolated task retention. On MT5, several baselines retain partial single-skill success while pairwise reliability drops sharply, consistent with skill-coupling collapse. Sleeping Robots mitigates this failure by separating online acquisition from offline consolidation and using frozen skill memories to rebalance the shared policy after each wake phase. The ablations indicate that the gains come from the consolidation mechanism, especially Nash-bargained sleep updates, rather than memory storage alone. Across Meta-World MT5 and SurgicAI, the same principle applies across two substrates: critic memories for state-based reinforcement learning and actor memories for vision-conditioned imitation. The results show that average success, backward transfer, pairwise reliability, and trajectory-structure preservation capture complementary aspects of shared-policy learning. Sleeping Robots is most relevant when a robot must retain one deployable policy, prior environments or trajectories are unavailable, and offline periods are available for consolidation.

\textbf{Limitations:} Frozen memories are local surrogates for learned behaviours. Critic memories can extrapolate when the current policy leaves the wake-time action distribution, while actor memories can become less informative after substantial policy drift. This creates a recovery-conservatism trade-off: stronger sleep updates may improve recovery but can over-trust memories outside their reliable region. We mitigate this with stored state or observation buffers, gradient normalization, bounded sleep updates, early stopping, and adaptive anchoring. Future work should explore uncertainty-aware memories, conservative critic objectives, adaptive sleep schedules, on-policy buffer regeneration, and trajectory-level distillation. The evaluation is also limited to two primary five-skill streams, one extended MT8 stream, original/reverse MT5 skill orders, and PRS over pre-registered skill pairs.

\bibliographystyle{IEEEtranN}
\bibliography{References}  

\clearpage
\appendix

\section{Implementation Details}
\label{app:implementation}

\subsection{Meta-World MT5}
\label{app:mt5_details}

\noindent\textbf{Full procedure.}
Algorithm~\ref{alg:sleeping_robots} summarizes the complete wake--sleep
lifecycle used in both benchmarks. The algorithm first performs wake-phase
training on the current skill, stores the resulting frozen skill memory, and
then performs offline sleep consolidation over all stored memories when
$k \geq 2$. The benchmark-specific details below specify the wake optimizer,
memory type, sleep hyperparameters, and evaluation protocol for MT5 and SurgicAI.

\begin{wrapfigure}[19]{r}{0.50\textwidth}
\vspace{-1.0em}
\centering
\begin{minipage}{0.48\textwidth}
\footnotesize
\hrule
\vspace{0.25em}
\captionof{algorithm}{Sleeping Robots}
\label{alg:sleeping_robots}
\hrule
\vspace{0.25em}
\begin{algorithmic}[1]
\Require Skills $\mathcal{T}_{1:K}$, policy $\pi_\theta$
\State $\mathcal{S}\gets\emptyset$, $\theta_0^{\mathrm{sleep}}\gets\theta$
\For{$k=1,\ldots,K$}
    \State Train $\pi_\theta$ on $\mathcal{T}_k$ using the substrate-specific wake objective
    \State Save $\theta_k^{\mathrm{wake}}\gets\theta$ and store $\mathcal{M}_k$
    \State $\mathcal{S}\gets\mathcal{S}\cup\{\mathcal{M}_k\}$
    \If{$k\geq2$}
        \State Set $\lambda_k$ by Eq.~\eqref{eq:lambda_anchoring}
        \For{$n=1,\ldots,N_{\mathrm{sleep}}$}
            \State Compute $\{g_i\}_{i=1}^k$ from $\mathcal{S}$
            \State Compute Nash direction $g^\star$
            \State $d\gets g^\star+\lambda_k(\theta-\theta_{k-1}^{\mathrm{sleep}})$
            \State $\theta\gets\mathrm{Adam}(\theta,d,\eta_{\mathrm{sleep}})$
            \State Update $\mathbf{e}$ if used
            \State Stop if sleep objective plateaus
        \EndFor
    \EndIf
    \State $\theta_k^{\mathrm{sleep}}\gets\theta$
\EndFor
\end{algorithmic}
\vspace{0.25em}
\hrule
\end{minipage}
\end{wrapfigure}

\noindent\textbf{Environment.}
We use Meta-World v2~\citep{yu2020meta} with the Sawyer robot arm.
The observation space is 39-dimensional and includes end-effector state,
gripper state, object state, and goal information. The action space is
4-dimensional and controls end-effector motion and gripper actuation. For
each skill, we use 50 randomized task configurations and sample one
configuration uniformly at the start of each episode.

\noindent\textbf{Skill stream.}
The MT5 stream consists of five skills learned in a fixed order:
\textsc{Reach} $\rightarrow$ \textsc{Door-Open} $\rightarrow$
\textsc{Drawer-Open} $\rightarrow$ \textsc{Window-Close} $\rightarrow$
\textsc{Button-Press-Topdown}. The skill order is fixed before training and
is shared by all methods.

\noindent\textbf{Policy and critic architecture.}
All MT5 methods use the same SAC policy substrate. The actor is a two-layer
MLP with 256 hidden units per layer and ReLU activations. For Sleeping
Robots, each hidden neuron additionally carries the homeostatic excitability
variable described in Section~\ref{sec:method}. The actor output
parameterizes a diagonal Gaussian policy over the continuous action space.
The log-standard-deviation bias is initialized to $-1.0$. The critic uses twin
Q-networks, each implemented as a two-layer MLP with 256 hidden units, ReLU
activations, and LayerNorm.

\noindent\textbf{SAC training.}
Wake-phase skill learning uses Soft Actor-Critic~\citep{haarnoja2018soft}
with automatic entropy tuning. Table~\ref{tab:sac_hyper} summarizes the
training hyperparameters used for MT5.

\begin{table}[h]
\vspace{-10pt}
\centering
\small
\caption{SAC hyperparameters for Meta-World MT5 wake training.}
\label{tab:sac_hyper}
\begin{tabular}{ll}
\toprule
\textbf{Setting} & \textbf{Value} \\
\midrule
Actor/critic optimizer & Adam, learning rate $3 \times 10^{-4}$ \\
Entropy coefficient & Automatically tuned \\
Entropy optimizer & Adam, learning rate $3 \times 10^{-4}$ \\
$\log \alpha$ clamp & $[-10, 4]$ \\
Target entropy & $-2.0$ \\
Reward scale & $1.0$ \\
Replay buffer size & $1{,}000{,}000$ \\
Update-to-data ratio & $5$ \\
Batch size & $256$ \\
Discount factor $\gamma$ & $0.99$ \\
Target-network update rate $\tau$ & $0.005$ \\
Random warmup & $5000$ environment steps per skill \\
Terminal handling & Time-limit truncations are not treated as terminal \\
\bottomrule
\end{tabular}
\end{table}

\noindent\textbf{Wake budgets and early stopping.}
Each skill is trained with a fixed maximum episode budget and a
skill-specific minimum budget before early stopping is allowed. Wake training
stops when the 100-episode rolling mean reward fails to improve by more than
2\% over the corresponding patience window. The budgets are given in
Table~\ref{tab:wake_budgets}.

\begin{table}[h]
\vspace{-10pt}
\centering
\small
\caption{Wake-phase budgets and early-stopping configuration for MT5.}
\label{tab:wake_budgets}
\begin{tabular}{lccc}
\toprule
\textbf{Skill} & \textbf{Maximum episodes} & \textbf{Minimum episodes} & \textbf{Patience} \\
\midrule
\textsc{Reach}        & 1500 &  500 &  500 \\
\textsc{Door-Open}    & 4000 & 1000 & 1000 \\
\textsc{Drawer-Open}  & 3000 & 1000 & 1000 \\
\textsc{Window-Close} & 2000 &  500 &  500 \\
\textsc{Button-Press} & 2500 &  800 &  800 \\
\bottomrule
\end{tabular}
\end{table}

\noindent\textbf{Homeostatic excitability.}
For Sleeping Robots, hidden-layer excitability variables are constrained to
$e_i \in [0.1, 5.0]$ and are updated only through the local homeostatic rule; they are not optimized by
backpropagation. The excitability hyperparameters are listed in
Table~\ref{tab:exc_hyper}.

\begin{table}[h]
\vspace{-10pt}
\centering
\small
\caption{Homeostatic excitability hyperparameters for MT5.}
\label{tab:exc_hyper}
\begin{tabular}{ll}
\toprule
\textbf{Setting} & \textbf{Value} \\
\midrule
Excitability bounds $[e_{\min}, e_{\max}]$ & $[0.1, 5.0]$ \\
Target activation $\alpha_{\mathrm{target}}$ & $0.10$ \\
Wake excitability rate $\eta_e^{\mathrm{wake}}$ & $1 \times 10^{-3}$ \\
Sleep excitability rate $\eta_e^{\mathrm{sleep}}$ & $1 \times 10^{-2}$ \\
Activation EMA factor $\beta$ & $0.01$ \\
Initial-skill novelty rate & $5 \times 10^{-3}$ for the first 500 episodes \\
\bottomrule
\end{tabular}
\end{table}

\noindent\textbf{Sleep consolidation.}
After each wake phase with $k \geq 2$, Sleeping Robots performs offline sleep
consolidation using the frozen memories accumulated so far. Sleep is skipped
for $k=1$ because only one memory is available and no cross-skill bargaining is
needed. For each skill, the sleep buffer contains 1000 raw states sampled
uniformly from the final 20\% of wake episodes. During sleep, the stored
normalizer snapshot $\nu_k$ is used when evaluating the corresponding frozen
critic, while the current actor uses its shared actor-side normalization. Sleep
uses Nash bargaining with warm-started Frank--Wolfe iterations over normalized
per-skill gradients. Hyperparameters are listed in Table~\ref{tab:sleep_hyper}.

\begin{table}[h]
\vspace{-10pt}
\centering
\small
\caption{Sleep-consolidation hyperparameters for MT5.}
\label{tab:sleep_hyper}
\begin{tabular}{ll}
\toprule
\textbf{Setting} & \textbf{Value} \\
\midrule
Sleep optimizer & Adam \\
Sleep learning rate $\eta_{\mathrm{sleep}}$ & $5 \times 10^{-4}$ \\
Maximum sleep iterations $N_{\mathrm{sleep}}$ & $4000$ \\
Early-stopping tolerance $\delta$ & $10^{-3}$ over a 100-iteration window \\
Gradient norm constant $\epsilon$ & $10^{-5}$ \\
Frank--Wolfe iterations & $50$ \\
Gradient normalization & Per-skill $\ell_2$ normalization \\
Frank--Wolfe warm start & Enabled \\
State buffer size $|\mathcal{B}_k|$ & $1000$ per skill \\
Sleep batch size & $64$ \\
\bottomrule
\end{tabular}
\end{table}

\noindent\textbf{Adaptive sleep anchoring.}
The anchoring diagnostic computes the distance
$c_i=\|\theta-\theta_i^{\mathrm{wake}}\|_2$ to each previous wake-end
snapshot, for $i<k$. If $\min_{i<k} c_i < \tau$ and $k<K$, the proximal
anchor is activated with strength
$\lambda=1\times10^{-2}$ toward the previous sleep-consolidated checkpoint
$\theta_{k-1}^{\mathrm{sleep}}$. We use $\tau=150$ for the MT5 experiments.
The anchor is not applied after the final skill.

\noindent\textbf{Frozen critic validation and diagnostics.}
Before freezing a skill critic, we evaluate it on a held-out batch from the
corresponding wake replay buffer. If the Bellman residual does not meet the
validation threshold, additional critic updates are performed before the memory
is stored. During sleep, we monitor implementation diagnostics on the stored MT5
state buffers, including frozen-critic value changes, Nash-gradient norms, and
early-stopping iterations. These quantities are used only to detect unusually
large or unstable sleep updates; they are not used as optimization targets or
performance metrics. The sleep optimizer instead relies on the safeguards in
Section~\ref{sec:method}, including stored late-wake states, per-skill
normalizer snapshots, gradient normalization, a fixed sleep budget, early
stopping, and adaptive anchoring.

In the MT5 runs, these diagnostics did not reveal unstable sleep updates or
systematic behavioral collapse after consolidation. We observed one localized
increase in frozen-critic value after the final sleep phase, driven primarily by
the \textsc{Window-Close} memory. However, the corresponding evaluated rollout
success on \textsc{Window-Close} after sleep remained high ($0.83$), while the
remaining memories showed value changes comparable to earlier sleep phases. We
therefore treat frozen-critic value changes only as sanity-check diagnostics and
base all behavioral conclusions on rollout evaluation through AFSR, PRS, BWT,
and SPI.

\noindent\textbf{Memory cost.}
Each MT5 skill memory consists of a frozen twin critic, a state buffer of 1000
raw states, and a normalizer snapshot. We compute memory from the stored
float32 tensors used by each method's continual-learning mechanism, excluding
temporary training buffers and evaluation-only checkpoints. At $K=5$, Sleeping
Robots stores frozen critics, state buffers, and normalizer snapshots for all
skills, for a total of approximately $3.7$ MB. This memory is larger than
state-only methods such as A-GEM, but remains within the same small-memory
regime as the other baselines and does not include previous actions, rewards,
transitions, or full trajectories.

\noindent\textbf{Evaluation protocol.}
After each skill, we evaluate the shared policy on all skills observed so far.
For MT5, each skill is evaluated using $N=100$ episodes with the same
task-randomization protocol used during training. Final results are reported
after the complete five-skill stream. For methods evaluated across multiple
seeds, we report mean $\pm$ standard error; for single-seed ablations, we
report point estimates. The MT5 PRS proxy uses the following six
pre-registered skill pairs:
\textsc{Reach}--\textsc{Door-Open},
\textsc{Door-Open}--\textsc{Drawer-Open},
\textsc{Reach}--\textsc{Drawer-Open},
\textsc{Door-Open}--\textsc{Window-Close},
\textsc{Drawer-Open}--\textsc{Button-Press}, and
\textsc{Window-Close}--\textsc{Button-Press}. PRS is evaluated only for
$k\geq2$.

\subsection{SurgicAI}
\label{app:surgicai_details}

\noindent\textbf{Environment.}
We use the SurgicAI benchmark~\citep{wu2024surgicai}, which is built on the
AMBF physics simulator for the da Vinci Research Kit (dVRK). The environment
contains two patient-side manipulators operating on a simplified suturing
phantom. We use the SurgicAI simulator configuration with the simple phantom,
ghost objects, and both PSM arms with actuators enabled. Each episode terminates
after 300 steps or when the subtask-specific success criterion is satisfied.

\noindent\textbf{Skill stream.}
The SurgicAI stream consists of five surgical subtasks learned in a fixed order:
\textsc{Approach} $\rightarrow$ \textsc{Place} $\rightarrow$
\textsc{Insert} $\rightarrow$ \textsc{Regrasp} $\rightarrow$
\textsc{Pullout}. These subtasks form a simplified suturing workflow, where the
policy must approach the entry point, orient and insert the needle, transfer the
needle, and pull it through the exit point. The skill order is fixed before
training and is shared by all methods.

\noindent\textbf{Observation and action spaces.}
Each policy receives a wrist-camera RGB image together with proprioceptive
state. The wrist camera is attached to the PSM2 tool-yaw link and provides
$640 \times 480$ RGB observations. We use the wrist camera as the visual input
for all SurgicAI methods. The low-dimensional state contains the achieved pose,
achieved-goal vector, and desired-goal vector. Actions are 7-dimensional
incremental commands over Cartesian pose and jaw motion, scaled by the SurgicAI
control step sizes.

\noindent\textbf{Demonstrations.}
Each subtask uses 100 expert demonstrations collected from prepositioned-motion
expert traces. Demonstrations include robot state, action, and synchronized
wrist-camera frames. All methods use the same demonstration set, skill order,
and evaluation protocol.

\noindent\textbf{Policy architecture.}
All SurgicAI methods use the same vision-conditioned ACT policy
substrate. The policy uses a ResNet-18 visual encoder and
an Action Chunking Transformer with transformer encoder--decoder blocks, learned
positional embeddings, a CVAE latent dimension of 32, and action chunk size 16.
The KL coefficient is set to 10. Transformer-based imitation policies use the
frozen-memory and sleep-consolidation components of Sleeping Robots, but do not
use the homeostatic excitability mechanism, which is applied only to
feedforward RL actors.

\noindent\textbf{Wake training.}
Wake-phase learning uses supervised behavior cloning on the current subtask.
Each skill is trained for up to 100 epochs with AdamW, using learning rate
$10^{-4}$ for the transformer, learning rate $10^{-5}$ for the visual backbone,
weight decay $10^{-4}$, and batch size 8. We use early stopping based on an
$\ell_1$ validation-loss plateau after a minimum of 30 epochs, and restore the
best-$\ell_1$ checkpoint at the end of wake training.

For Sleeping Robots, wake training on skill $k \geq 2$ also includes
output-space distillation from the frozen actor memories of previous skills:
\begin{equation}
\mathcal{L}_{\mathrm{wake}}
=
\mathcal{L}_{\mathrm{BC}}
+
\gamma_k
\sum_{i<k}
\mathbb{E}_{o \sim \mathcal{B}_i}
\left[
\left\|
\pi_\theta(o) - \pi_{\theta_i^{\mathrm{wake}}}(o)
\right\|_2^2
\right].
\end{equation}
We use $\gamma_k=1.0$ for $k=2$ and $\gamma_k=0.5$ for $k\geq3$ to balance
new-skill acquisition against preservation of previous actor outputs. The same
frozen actor memories are later used during sleep.

\begin{table}[h]
\vspace{-10pt}
\centering
\small
\caption{ACT wake-training hyperparameters for SurgicAI.}
\label{tab:surgicai_wake_hyper}
\begin{tabular}{ll}
\toprule
\textbf{Setting} & \textbf{Value} \\
\midrule
Policy substrate & ACT with ResNet-18 visual encoder \\
Transformer learning rate & $1 \times 10^{-4}$ \\
Backbone learning rate & $1 \times 10^{-5}$ \\
Optimizer & AdamW \\
Weight decay & $1 \times 10^{-4}$ \\
Batch size & $8$ \\
Maximum wake epochs & $100$ per skill \\
Minimum epochs before early stopping & $30$ \\
Early stopping criterion & $\ell_1$ validation-loss plateau \\
Checkpoint selection & Best-$\ell_1$ wake checkpoint \\
Action chunk size & $16$ \\
CVAE latent dimension & $32$ \\
KL weight & $10$ \\
Wake distillation weight $\gamma_k$ & $1.0$ for $k=2$, $0.5$ for $k\geq3$ \\
\bottomrule
\end{tabular}
\end{table}

\noindent\textbf{Sleep consolidation.}
After each wake phase with $k \geq 2$, Sleeping Robots performs offline sleep
consolidation over the frozen actor memories accumulated so far. Each memory
contains a frozen wake-end actor and an observation buffer. The sleep objective
uses actor-snapshot distillation on stored observations:
\begin{equation}
\mathcal{L}^{\pi}_i(\theta)
=
\mathbb{E}_{o \sim \mathcal{B}_i}
\left[
\left\|
\pi_\theta(o) - \pi_{\theta_i^{\mathrm{wake}}}(o)
\right\|_2^2
\right].
\end{equation}
Per-skill gradients are combined using the same Nash-bargained consolidation
procedure described in Section~\ref{sec:method}. Sleep uses Adam with
learning rate $10^{-5}$, sleep batch size 8, at most 1000 sleep iterations, and
early stopping when the sleep objective plateaus. The proximal sleep anchor uses
strength $\lambda=10^{-2}$ toward the previous sleep-consolidated checkpoint.
Each skill memory stores 1024 observation samples for sleep.

\begin{table}[h]
\vspace{-10pt}
\centering
\small
\caption{Sleep-consolidation hyperparameters for SurgicAI.}
\label{tab:surgicai_sleep_hyper}
\begin{tabular}{ll}
\toprule
\textbf{Setting} & \textbf{Value} \\
\midrule
Sleep optimizer & Adam \\
Sleep learning rate $\eta_{\mathrm{sleep}}$ & $1 \times 10^{-5}$ \\
Maximum sleep iterations $N_{\mathrm{sleep}}$ & $1000$ \\
Early-stopping tolerance & $10^{-3}$ objective-window change \\
Frank--Wolfe iterations & $5$ \\
Gradient normalization & Per-skill $\ell_2$ normalization \\
Observation buffer size $|\mathcal{B}_k|$ & $1024$ per skill \\
Sleep batch size & $8$ \\
Anchor strength $\lambda$ & $1 \times 10^{-2}$ \\
Memory type & Frozen actor snapshot + observation buffer \\
Excitability & Not used for transformer-based ACT policies \\
\bottomrule
\vspace{-15pt}
\end{tabular}
\end{table}

\noindent\textbf{Baselines.}
All SurgicAI baselines use the same ACT policy substrate, demonstrations, skill
order, and wake-training budget as Sleeping Robots. Sequential BC trains on each
new subtask without an explicit continual-learning or consolidation mechanism.
LwF uses wake-time output distillation from a rolling previous actor. EWC applies
a diagonal Fisher penalty to protect parameters from previous subtasks.
M2Distill extends distillation to include intermediate visual features in
addition to action outputs. SR Wake-Only is an ablation of our method that keeps
wake-time multi-anchor distillation and frozen actor memories but removes the
offline sleep phase. MT-Joint trains a single ACT policy on the union of all five
subtask demonstration sets and is reported only as an oracle reference, not as a
continual-learning method.

\noindent\textbf{Evaluation protocol.}
After each skill, we evaluate the shared policy on every subtask observed so far.
For SurgicAI, each subtask is evaluated using $N=20$ episodes per skill in the
main results. Final results are reported after the complete five-skill stream.
AFSR is computed as the average success over learned subtasks. PRS is computed
using the pairwise-min protocol over seven pre-registered workflow-ordered
subtask pairs:
\textsc{Approach}--\textsc{Place},
\textsc{Place}--\textsc{Insert},
\textsc{Insert}--\textsc{Regrasp},
\textsc{Regrasp}--\textsc{Pullout},
\textsc{Approach}--\textsc{Insert},
\textsc{Place}--\textsc{Regrasp}, and
\textsc{Insert}--\textsc{Pullout}.
BWT is computed from the change in earlier-skill success after learning later subtasks. SPI measures preservation of the top-three PCA directions of the
reference observation-space trajectory.

\noindent\textbf{Compute.}
SurgicAI training and evaluation are more expensive than MT5 because each policy
uses a vision-conditioned ACT model and rollouts are performed in AMBF. Training
and evaluation are run on a single GPU. Due to simulator cost, SurgicAI
evaluations use fewer episodes per skill than MT5, as described in
Section~\ref{sec:experiments}.

\section{Baseline Details}
\label{app:baseline_details}

\subsection{Meta-World MT5 Baselines}
\label{app:mt5_baselines}

All MT5 baselines use the same SAC backbone, actor architecture, skill order,
wake-training budget, and evaluation protocol as Sleeping Robots unless stated
otherwise. Each method is adapted to the sequential RL setting as follows.

\noindent\textbf{EWC~\citep{kirkpatrick2017overcoming}.}
After each skill, we store a parameter snapshot $\theta_k^\star$ and diagonal
Fisher estimate $F_k$. During later wake phases, the actor loss includes
\begin{equation}
\frac{\lambda}{2}
\sum_{i<k}
F_i(\theta-\theta_i^\star)^2 .
\end{equation}
We use $\lambda=1$ after tuning over $\{500,100,1\}$. Fisher estimates are
computed from 2000 states per skill and normalized to unit Frobenius norm.

\noindent\textbf{A-GEM~\citep{chaudhry2018efficient}.}
We store 1000 raw states per past skill. During wake training, a reference
gradient is recomputed every 30 update-to-data steps using the corresponding
frozen critic. Conflicting actor gradients are projected onto the feasible
half-space before the update.

\noindent\textbf{DER~\citep{buzzega2020dark}.}
For each skill, we store 1000 states and the actor's Gaussian mean and
log-standard-deviation at storage time. During later wake phases, we add
output-space distillation to the actor loss as below with $\alpha_{\mathrm{DER}}=0.1$.
\begin{equation}
\alpha_{\mathrm{DER}}
\|\mu_\theta(s)-\mu_{\mathrm{stored}}(s)\|_2^2 ,
\end{equation}

\noindent\textbf{CLEAR~\citep{rolnick2019experience}.}
CLEAR stores 1000 transitions $(s,a,r,s',\mathrm{done})$ per previous skill
with behavioral-cloning targets. Each SAC update uses a 50/50 mixture of
current-skill and replayed samples, with a BC loss coefficient
$\alpha_{\mathrm{BC}}=0.01$. V-trace is disabled because SAC is off-policy.

\noindent\textbf{Reset \& Distill~\citep{ahn2025reset}.}
For each skill, the actor and critic are reset and trained from scratch. The
trained actor is frozen as an expert. After each skill, all experts are
distilled into a shared actor for 2000 iterations using batch size 256 and an
MSE loss on Gaussian mean and log-standard-deviation, with 1000 states per
expert.

\noindent\textbf{Progress \& Compress~\citep{schwarz2018progress}.}
An active column learns the current skill and is then compressed into a shared
knowledge base. We use MSE distillation and EWC regularization with
Frobenius-normalized Fisher estimates. The active column is reset at each skill.

\noindent\textbf{Nash-MTL~\citep{navon2022multi} adapted to sequential RL.}
We replace simultaneous task losses with frozen-critic surrogate gradients from
stored skill memories. During wake training, per-skill gradients are normalized
and combined with the same Frank--Wolfe Nash solver used by Sleeping Robots.
Unlike Sleeping Robots, bargaining occurs during wake rather than offline sleep.

\noindent\textbf{PCGrad~\citep{yu2020gradient} adapted to sequential RL.}
PCGrad uses the same frozen-critic surrogate gradients as adapted Nash-MTL.
Conflicting normalized gradients are projected orthogonally before the actor
update, and the projected update is rescaled to the original gradient norm.

\noindent\textbf{CompoNet~\citep{malagon2025self} adapted to SAC.}
Each skill adds an actor head and twin critic. A shared feature trunk is frozen
after the first skill, and module outputs are combined by softmax attention.
Attention is initialized with self-bias $\alpha_{\mathrm{self}}=2.0$.

\noindent\textbf{MT-joint oracle.}
The oracle trains multi-task SAC on all five MT5 skills simultaneously with
uniform task sampling for 13{,}000 total episodes. It is not a
continual-learning method and is reported only as an upper-bound reference; BWT
is zero by definition. After training, the oracle achieves
AFSR\,=\,0.975\,$\pm$\,0.007 and PRS\,=\,0.958\,$\pm$\,0.011 over three
evaluation seeds with $N=100$ episodes per skill. This confirms that high
single-skill success and high pairwise reliability are achievable on the MT5
stream when all skills are available jointly, and contextualizes the gap faced
by sequential methods.

\subsection{Memory Budget}
\label{app:memory_budget}

Table~\ref{tab:memory_budget} summarizes the method-specific memory used by
each MT5 method. Our evaluation pipeline saves frozen critics and state buffers
for diagnostic and evaluation purposes, but these objects are counted as
continual-learning memory only for methods that explicitly use them during their
update. The table therefore reports the storage used by each method's own
continual-learning mechanism. The goal is not exact byte-level matching, since
different methods store different objects, but to clarify that all methods
operate in a small-memory regime.

\begin{table}[h]
\vspace{-10pt}
\centering
\small
\caption{Method-specific memory stored in the MT5 experiments at $K=5$.
\ding{51} indicates that the information is stored by the continual-learning
mechanism, and \ding{55} indicates that it is not stored.}
\label{tab:memory_budget}
\begin{tabular}{lccccr}
\toprule
\textbf{Method}
& \textbf{States}
& \textbf{Actions}
& \textbf{Rewards/}
& \textbf{CL-specific storage}
& \textbf{Size} \\
& & & \textbf{trans.} & & \textbf{(MB)} \\
\midrule
EWC
& \ding{55} & \ding{55} & \ding{55}
& Fisher diagonal + $\theta^\star$ per skill
& 3.0 \\

A-GEM
& \ding{51} & \ding{55} & \ding{55}
& 1000 states per skill
& 0.7 \\

DER
& \ding{51} & \ding{55} & \ding{55}
& 1000 states + policy outputs per skill
& 1.6 \\

CLEAR
& \ding{51} & \ding{51} & \ding{51}
& 1000 $(s,a,r,s')$ + behavior targets per skill
& 2.5 \\

R\&D
& \ding{51} & \ding{55} & \ding{55}
& Frozen expert actor + states per skill
& 2.2 \\

P\&C
& \ding{55} & \ding{55} & \ding{55}
& Knowledge base + Fisher diagonal
& 3.0 \\

Nash-MTL
& \ding{51} & \ding{55} & \ding{55}
& Frozen critics + states + normalizer snapshots
& 3.7 \\

PCGrad
& \ding{51} & \ding{55} & \ding{55}
& Frozen critics + states + normalizer snapshots
& 3.7 \\

CompoNet
& \ding{55} & \ding{55} & \ding{55}
& Frozen skill modules
& 4.0 \\

\midrule
\textbf{Sleeping Robots}
& \ding{51} & \ding{55} & \ding{55}
& Frozen critics + states + normalizer snapshots
& \textbf{3.7} \\
\bottomrule
\end{tabular}
\vspace{-10pt}
\end{table}

Sleeping Robots stores unordered state buffers, frozen skill critics, and
normalizer snapshots, but does not store previous actions, rewards, transitions,
or full trajectories. Although Sleeping Robots is not the smallest-memory method,
all methods remain within a comparable small-memory regime. Nash-MTL and PCGrad
use the same frozen-critic memory interface as Sleeping Robots, so comparisons
against these baselines isolate the consolidation mechanism rather than the
stored information. The relevant distinction is therefore not storage capacity
alone, but how the stored information is used: replay-based methods reuse
transition-level data, whereas Sleeping Robots uses frozen skill memories for
offline consolidation.

\section{Additional Experiments}
\label{app:additional_experiments}

\subsection{Robustness to Skill Order and Stream Length}
\label{app:order_length}

We include two additional Meta-World robustness checks using the same SAC substrate, frozen-critic memory interface, and sleep-consolidation procedure as the main MT5 experiment. First, we evaluate a reverse-order MT5 stream, in which the original order
\textsc{Reach} $\rightarrow$ \textsc{Door-Open} $\rightarrow$ \textsc{Drawer-Open} $\rightarrow$ \textsc{Window-Close} $\rightarrow$ \textsc{Button-Press}
is reversed while keeping the same five skills and evaluation protocol. This isolates sensitivity to acquisition order. Second, we evaluate an extended MT8 stream by continuing from the canonical MT5 checkpoint with three additional Meta-World tasks,
\textsc{Window-Open}, \textsc{Drawer-Close}, and \textsc{Handle-Press}. This tests sensitivity to a longer sequential stream.

Final-policy metrics are reported in Table~\ref{tab:order_length_sensitivity} using $N=100$ rollouts per skill. For the original and reverse MT5 streams, PRS uses the same six pre-registered skill pairs as the main experiment, indexed by skill identity rather than stream position. For MT8, PRS is computed over all $\binom{8}{2}=28$ skill pairs, so PRS values should be interpreted as a stream-level robustness check rather than a one-to-one comparison with the MT5 pair set. SPI uses the same top-three PCA alignment protocol as the main results. The canonical MT5 result is averaged over three seeds, while the reverse MT5 and MT8 checks are single-seed.

\end{document}